\documentclass[sigconf,nonacm]{acmart}

\usepackage{amsmath}
\usepackage{booktabs}
\usepackage{graphicx}
\usepackage[most]{tcolorbox}
\microtypesetup{expansion=false}
\graphicspath{{fig/}}

\setcopyright{none}
\acmDOI{}
\acmISBN{}

\title{SKALOS: Learning to Distinguish Similar Skills with Capability Graphs}

\author{Wenhan Xiao}
\affiliation{
  \institution{Beihang University}
  \city{Beijing}
  \country{China}
}
\email{xiaowenhan@buaa.edu.cn}

\author{Ziwei Zhang}
\authornote{Corresponding author.}
\affiliation{
  \institution{Beihang University}
  \city{Beijing}
  \country{China}
}
\email{zwzhang@buaa.edu.cn}

\author{Xingcheng Fu}
\affiliation{
  \institution{Guangxi Normal University}
  \city{Guilin}
  \country{China}
}
\email{fuxc@gxnu.edu.cn}

\author{Chuanyue Yu}
\affiliation{
  \institution{Nankai University}
  \city{Tianjin}
  \country{China}
}
\email{yuchuanyue@mail.nankai.edu.cn}

\author{Haonan Yuan}
\affiliation{
  \institution{Tsinghua University}
  \city{Beijing}
  \country{China}
}
\email{yuanhn@mail.tsinghua.edu.cn}

\author{Qingyun Sun}
\affiliation{
  \institution{Beihang University}
  \city{Beijing}
  \country{China}
}
\email{sunqy@buaa.edu.cn}

\author{Runhua Xu}
\affiliation{
  \institution{Beihang University}
  \city{Beijing}
  \country{China}
}
\email{runhua@buaa.edu.cn}

\author{Jianxin Li}
\affiliation{
  \institution{Beihang University}
  \city{Beijing}
  \country{China}
}
\email{lijx@buaa.edu.cn}

\renewcommand{\shortauthors}{Xiao et al.}
\makeatletter
\def\@mkauthors@iii{%
  \gdef\@currentauthors{}%
  \gdef\@currentaffiliation{}%
  \global\setbox\mktitle@bx=\vbox{%
    \unvbox\mktitle@bx\par\medskip
    \begingroup
    \hsize=\textwidth
    \centering
    {\large
      Wenhan Xiao\textsuperscript{1}\quad
      Ziwei Zhang\textsuperscript{1,*}\quad
      Xingcheng Fu\textsuperscript{2}\quad
      Chuanyue Yu\textsuperscript{3}\par
      Haonan Yuan\textsuperscript{4}\quad
      Qingyun Sun\textsuperscript{1}\quad
      Runhua Xu\textsuperscript{1}\quad
      Jianxin Li\textsuperscript{1}\par}
    \smallskip
    {\small
      \textsuperscript{1}Beihang University, Beijing, China\quad
      \textsuperscript{2}Guangxi Normal University, Guilin, China\par
      \textsuperscript{3}Nankai University, Tianjin, China\quad
      \textsuperscript{4}Tsinghua University, Beijing, China\par}
    \smallskip
    {\small\ttfamily
      \{xiaowenhan, zwzhang, sunqy, runhua, lijx\}@buaa.edu.cn\par
      \href{mailto:fuxc@gxnu.edu.cn}{fuxc@gxnu.edu.cn}\quad
      \href{mailto:yuchuanyue@mail.nankai.edu.cn}{yuchuanyue@mail.nankai.edu.cn}\par
      \href{mailto:yuanhn@mail.tsinghua.edu.cn}{yuanhn@mail.tsinghua.edu.cn}\par}
    \endgroup
    \bigskip
  }%
}
\makeatother
\hypersetup{pdfauthor={Wenhan Xiao, Ziwei Zhang, Xingcheng Fu, Chuanyue Yu, Haonan Yuan, Qingyun Sun, Runhua Xu, Jianxin Li}}

\begin{document}

\begin{abstract}
Skills provide LLM agents with reusable procedural knowledge, making appropriate skill selection essential for reliable agent behavior.
As skill libraries expand, distinguishing semantically similar candidates becomes increasingly difficult.
Existing methods encode or rewrite skill descriptions for relevance-based ranking, yet heterogeneous descriptions can obscure capability differences.
To address this challenge, we propose \textbf{SKALOS}, a graph-native reranker that compares skills via explicit capability evidence. Its multi-hierarchy graph aligns equivalent expressions across skills while preserving skill-specific capability evidence. Since individually relevant capability elements may not jointly support a query, semiring-based capability witness learning scores each branch by its weakest evidence, checks joint support for the extracted requirements within one implementation, and selects the strongest alternative. To reduce the influence of extraction errors, we develop ranking-feedback graph calibration that adjusts unreliable evidence for candidate comparison. Experiments on SkillRet show that SKALOS improves NDCG@10 by 2.68 percentage points over a matched text-only reranker. Witness-guided skill selection also increases ALFWorld task success rate from 22\% to 29\%.
\end{abstract}

\ccsdesc[500]{Information systems~Retrieval models and ranking}
\ccsdesc[300]{Mathematics of computing~Graph algorithms}
\keywords{Agent skills, Skill reranking, Capability graphs, LLM agents}

\maketitle

\section{Introduction}
\label{sec:introduction}

Web agents rely on reusable skills to turn user requests into executable workflows across heterogeneous tools and services. Selecting suitable skills requires matching their procedures, outputs, and usage constraints to the requested workflow, since similar descriptions can correspond to different execution capabilities~\cite{DBLP:conf/www/Cai00Z00C25,DBLP:conf/www/LiJJDJWWZWLD26,DBLP:conf/www/ShiGY0CCYVR25,DBLP:conf/www/ZhuangZZZH26}. Within this setting, skills provide reusable procedural knowledge that guides how agents use available tools and services. By retaining established methods for reuse across tasks, skill libraries support the accumulation of agent capabilities and reduce the need to derive procedures from scratch~\cite{wang2023voyager}. As these libraries expand, selecting appropriate skills becomes increasingly important. Recent advances in skill retrieval and reranking address this need through specialized models and richer skill descriptions, improving access to relevant candidates in large libraries~\cite{cho2026skillret,li2025skillflow,zheng2026skillrouterskillroutingllm}.

\begin{figure}[t]
    \centering
    \includegraphics[width=\columnwidth]{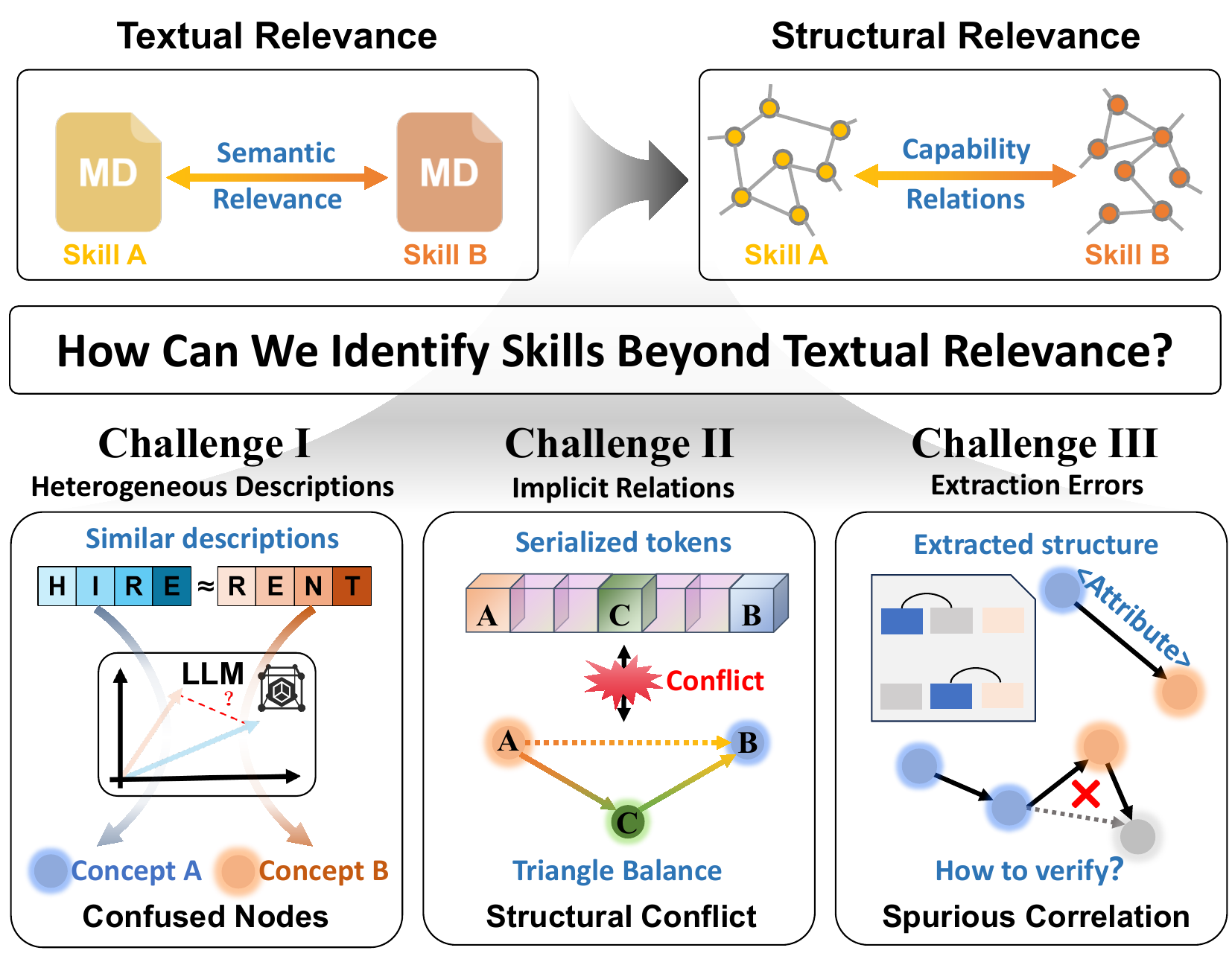}
    \caption{Three challenges in capability-aware skill selection: heterogeneous descriptions, implicit relations in text, and errors in extracted structures.}
    \Description{Equivalent expressions may receive different embeddings; serialized relations must be recovered from textual context; structure extraction may introduce missing or spurious relations.}
    \label{fig:challenges}
\end{figure}

Despite these advances, relevance alone is insufficient for selecting among semantically similar skills. Candidates may receive similar relevance scores even when their evidence supports different requirements through separate procedures, rather than jointly within one implementation. Heterogeneous descriptions also obscure shared capabilities and meaningful implementation differences. The central challenge is therefore not merely to retrieve related descriptions, but to establish a common basis for comparing the capability evidence of easily confused candidates.

Recent studies attempt to expose skill structure by extracting workflows, dependencies, and relations from skill documents~\cite{ding2026skill2query,liu2026graph,bai2026skilldag}. Recent skill graphs primarily organize dependencies and execution relationships across skills, clarifying how skills can be combined or invoked rather than how semantically similar candidates can be distinguished. Moreover, extracting structure does not by itself resolve how equivalent evidence should be compared, how its relations should be used, or how extraction errors should be handled. Figure~\ref{fig:challenges} illustrates three resulting challenges.

\noindent\textbf{Challenge I: Heterogeneous descriptions.}
Skills developed by different authors often express equivalent capabilities through different terminology and sentence structures. When encoded as text, equivalent evidence may receive different representations and ranking scores; prior work also shows that tool-selection preferences are sensitive to textual descriptions~\cite{faghih2025tool}. This motivates aligning equivalent phrases across skill documents~\cite{vashishth2018cesi,dash2021open}. However, alignment should not erase meaningful implementation differences: two skills may share a broad capability while differing in methods, outputs, or constraints that determine their suitability for a task. The challenge is to align equivalent capabilities while retaining the details that distinguish skill implementations.

\noindent\textbf{Challenge II: Implicit relations.}
A skill may mention all the operations and outputs relevant to a task, yet provide no single procedure that meets all of the task's requirements. Text-based representations encode these relations through word order and context, making their interpretation sensitive to wording, ordering, and templates~\cite{fatemi2024talk,sclar2024quantifying}. The presence of individually relevant elements therefore does not establish capability completeness: their supporting relations and implementation boundaries also matter. The challenge is to determine whether the extracted requirements are jointly supported within one implementation, without combining evidence from separate procedures.

\noindent\textbf{Challenge III: Extraction errors.}
Extracting and canonicalizing capability evidence from skill documents may introduce incorrect node mappings or unsupported relations~\cite{ye2022generative,dash2021open}. Incorrect mappings can obscure skill-specific distinctions, while unsupported relations can introduce misleading capability evidence. Calibrating such evidence is difficult because ranking labels identify preferred skills without specifying the contribution of each extracted node or relation. Moreover, evidence irrelevant to a particular query may still be correctly extracted. The challenge is to suppress unreliable structure while retaining useful capability evidence.

To address these challenges, we propose \textbf{SKALOS}, a graph-native reranker that uses textual relevance and structural support to compare candidate skills. First, our graph aligns shared capabilities and retains local evidence specific to each skill. Next, witness learning scores each branch by its weakest evidence, checks joint requirement coverage within one implementation, and selects the strongest valid alternative. Finally, graph calibration adjusts node representations and relation weights by ranking feedback obtained via element masking. Experiments on SkillRet demonstrate improved skill reranking across multiple rerankers and robustness to structural perturbations, and controlled ALFWorld tasks confirm the downstream value of witness-guided skill selection.

The witness tests whether one procedure jointly supports the query, rather than treating a union of relevant elements as complete evidence. Calibration then controls how strongly uncertain nodes and relations influence candidate ranking. These stages preserve the distinction between textual relevance, structural support, and evidence reliability throughout skill comparison.

Our contributions are summarized as follows:
\begin{itemize}
    \item We design a multi-hierarchy capability graph that makes heterogeneous skills comparable while preserving their distinctive implementation evidence.
    \item We propose a graph-native skill reranking framework that integrates relation-aware graph modeling with semiring-based capability witness learning.
    \item We develop ranking-feedback graph calibration to limit the influence of unreliable evidence during skill selection.
\end{itemize}

\section{Related Work}

\subsection{Text-Based Skill Retrieval and Reranking}

Recent work addresses large-scale skill selection through specialized text embeddings and retrieve-and-rerank pipelines, using skill descriptions to match queries with candidate skills~\cite{cho2026skillret,li2025skillflow,zheng2026skillrouterskillroutingllm}. Related tool-retrieval methods also exploit specialized retrieval models and library hierarchies to improve candidate selection~\cite{shi2025retrieval,zheng2024toolrerank}. Despite these advances, text-based matching leaves internal capability relations implicit in the input sequence. Equivalent evidence expressed through different descriptions may therefore receive different representations, consistent with findings that tool-selection preferences are sensitive to textual descriptions~\cite{faghih2025tool}. High semantic relevance alone does not explicitly establish whether a candidate jointly supports the required methods, artifacts, and constraints.

\subsection{Retrieval-Augmented Generation}

RAG retrieves and organizes external evidence for generation~\cite{sarthi2024raptor,gutierrez2024hipporag,he2024g,10.1145/3777378}. Retrieval augmentation also supports access to agent skills~\cite{su2026skill}, while skill graphs model how multiple skills can be combined~\cite{liu2026graph,bai2026skilldag,li2026caskg}. SKALOS tests whether one candidate implementation jointly supports the query requirements.

\subsection{Structured Skill Representation}

Existing structured approaches to skill selection operate at different levels. Inter-skill graph methods represent skills as nodes and encode dependencies, compatibility, and procedural relations to support retrieval and composition~\cite{liu2026graph,bai2026skilldag,li2026caskg}. Their structural information primarily describes how skills relate to one another and can be organized into task-relevant combinations. Our focus is the internal capability composition of individual candidates, where semantically similar skills may differ in the methods, artifacts, and constraints they can jointly support.

Other methods expose capability distinctions through discriminative text or retrieval supervision derived from structure~\cite{wang2026skills,ding2026skill2query}. They enrich textual representations or training signals with capability information. SKALOS instead uses capability relations directly in candidate scoring, making joint requirement support within one implementation an explicit ranking signal.

Complementary research on extraction and canonicalization supports relation recovery and alignment of heterogeneous expressions~\cite{ye2022generative,kalyan2020medical,dash2021open,vashishth2018cesi}. However, extraction errors and mistaken merges can distort the evidence used for candidate comparison.
The remaining challenge is to align capability evidence across candidates, retain implementation-specific distinctions, and assess joint support when the extracted structure is imperfect.

\section{Method}
\label{sec:method}

\begin{figure*}[t]
    \centering
    \includegraphics[width=\textwidth]{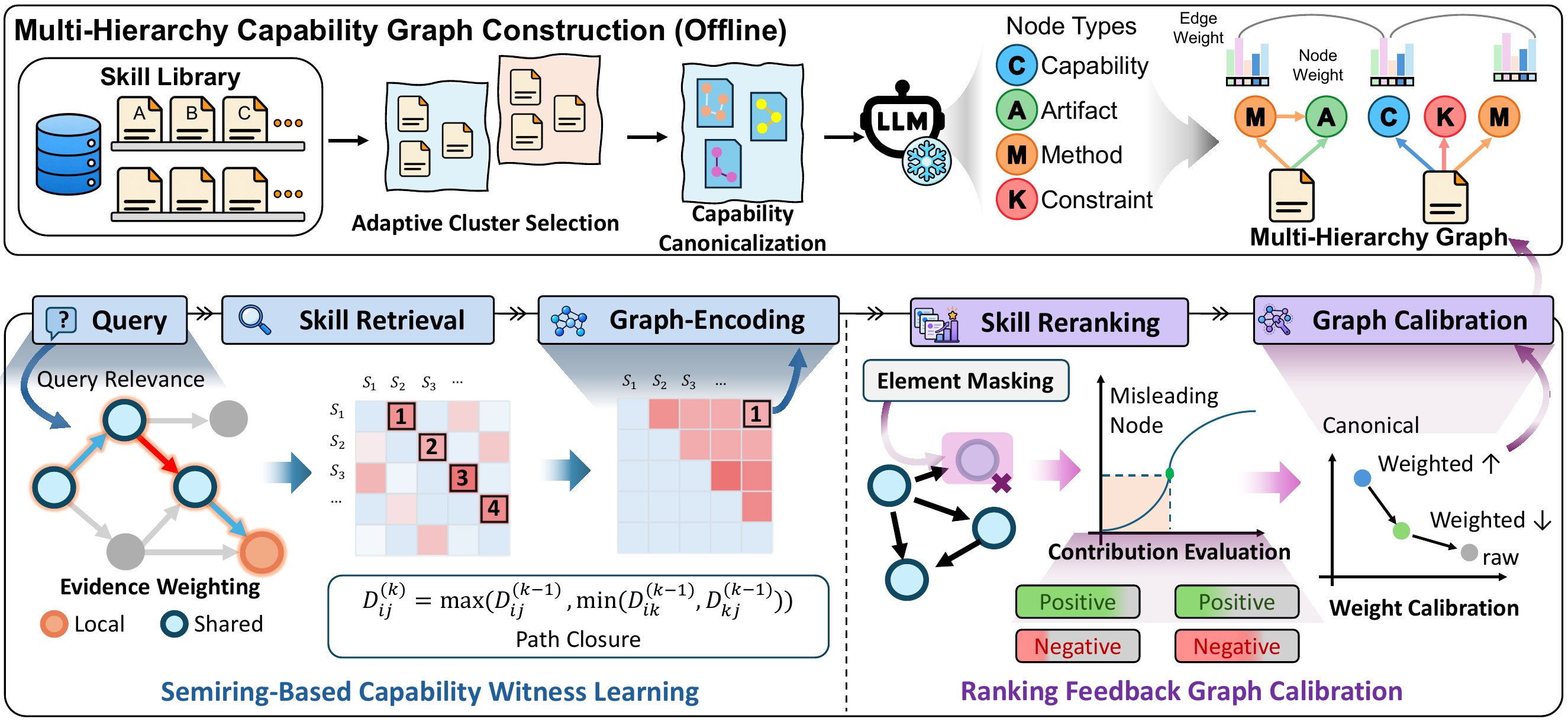}
    \caption{Overview of SKALOS. Multi-hierarchy capability graphs organize shared and skill-specific evidence; semiring-based witness learning aggregates requirement-covering branches anchored to one implementation; ranking-feedback graph calibration adjusts unreliable evidence used for skill reranking.}
    \Description{The framework constructs multi-hierarchy capability graphs from skill documents, encodes query-conditioned graph evidence, forms requirement-covering capability witnesses using a Max-Min semiring, and calibrates evidence confidence through ranking feedback.}
    \label{fig:framework}
\end{figure*}

Figure~\ref{fig:framework} presents the three stages of SKALOS. Graph construction aligns equivalent capability expressions and retains the evidence specific to each skill. Witness learning checks joint requirement coverage within one implementation, scores its branches by their weakest evidence, and selects the strongest alternative. Graph calibration uses ranking feedback to adjust unreliable evidence before scoring candidate skills against the query requirements.

\subsection{Multi-Hierarchy Capability Graphs}

\label{sec:graph-construction}

Our graph aligns equivalent capability expressions to compare similar skills, while preserving the methods, outputs, and constraints that distinguish individual implementations. Its shared and local layers retain both common references and skill-specific evidence. Figure~\ref{fig:capability-graph-example} illustrates the four node types and three typed relations extracted from a skill document.

\begin{figure}[htbp]
    \centering
\begin{tcolorbox}[
    enhanced,
    width=\linewidth,
    colback=white,
    colframe=black!80,
    colbacktitle=black!80,
    coltitle=white,
    rounded corners,
    boxrule=0.7pt,
    boxsep=2pt,
    left=2pt,right=2pt,top=3pt,bottom=3pt,
    toptitle=3pt,bottomtitle=3pt,
    fonttitle=\footnotesize\bfseries,
    title={\makebox[0.43\linewidth][l]{Original Skill}Extracted Graph},
    sidebyside,
    lefthand ratio=0.41,
    sidebyside gap=10pt,
    sidebyside align=top seam,
    fontupper=\footnotesize,
    fontlower=\footnotesize,
    before skip=6pt,after skip=6pt
]
\raggedright
\textbf{PDF Table Extraction}\par
Extract tables from scanned PDF documents.
\par\medskip
\textbf{Procedure}\par
Apply OCR to recognize table content and reconstruct rows and columns.
\par\medskip
\textbf{Output}\par
Export the extracted tables as CSV files.
\par\medskip
\textbf{Constraint}\par
The OCR procedure supports English text only.
\tcblower
\begin{lstlisting}[
    basicstyle=\fontsize{7}{9}\selectfont\ttfamily,
    columns=fullflexible,keepspaces=true,
    showstringspaces=false,
    aboveskip=0pt,belowskip=0pt,
    breaklines=false,numbers=none,frame=none
]
{
  "nodes": {
    "c":["capability","PDF tables"],
    "m":["method","OCR"],
    "a":["artifact","CSV"],
    "k":["constraint",
         "English only"]
  },
  "edges": [
    ["m", "implements", "c"],
    ["m", "produces", "a"],
    ["k", "constrains", "m"]
  ]
}
\end{lstlisting}
\end{tcolorbox}
\caption{A constructed skill graph with four node types and three typed relations. Its OCR method produces CSV tables under an English-only constraint.}
\label{fig:capability-graph-example}
\Description{A skill description on the left is represented by four typed nodes and their relations in JSON on the right.}
\end{figure}

\paragraph{\textbf{Capability canonicalization.}}
To align heterogeneous expressions within a finite canonical node bank, we perform \emph{cluster-scoped capability canonicalization}. Skills are first grouped in the embedding space, with the cluster count adaptively selected according to compactness, separation, and cluster size~\cite{rousseeuw1987silhouettes}. Within each cluster, we extract capability, method, artifact, and constraint expressions and consolidate semantically equivalent expressions of the same type~\cite{kalyan2020medical,dash2021open}. Canonicalizing expressions separately within each cluster and node type reduces the risk of merging expressions with different contextual meanings or semantic roles. These nodes provide a common basis for comparing differently worded skill descriptions.

\paragraph{\textbf{Shared and local evidence.}}
Canonicalization reveals both recurring nodes used by multiple skills and nodes specific to a single skill. Although recurring nodes expose common capabilities, skill-specific nodes may capture precisely the implementation details that distinguish otherwise similar candidates. Removing them would make these candidates less distinguishable. We place recurring nodes in a shared layer and skill-specific nodes in a local layer, preserving both common references and implementation details.

\paragraph{\textbf{Multi-hierarchy graph.}}
We organize the shared and local evidence in a multi-hierarchy graph. Skills may contain similar elements but connect them differently, making typed relations necessary for comparing their implementations.
We encode these connections through typed relations, including \textsc{requires}, \textsc{supports}, \textsc{implements}, \textsc{produces}, and \textsc{constrains}, making implementation relationships explicit for subsequent comparison. For a candidate skill $s$, we denote the resulting graph by $G_s=(V_s,E_s)$. Each typed edge has an inverse view with the same relation identity; $E_s^{\pm}$ denotes this augmented edge set.

\subsection{Semiring-Based Capability Witnesses}
\label{sec:witness}

A skill may contain several highly relevant elements without providing a jointly supported implementation. Additive aggregation can conceal a missing requirement behind other strong matches, while selecting a single strong path may bypass a necessary branch. Moreover, evidence belonging to different implementations should not be assembled into one apparently complete witness. We therefore introduce capability witness learning to require joint coverage within one implementation while retaining alternative implementations as competing sources of support.

Let $\mathcal{R}_q=\{r_1,\ldots,r_K\}$ denote the typed capability requirements expressed by query $q$, such as its required capability, method, artifact, and constraint. Requirement phrases are aligned to the canonical node bank by type-constrained query--node matching; types not specified by the query are not forced into the witness. The witness treats extracted requirements as mandatory conjunctions; optional, disjunctive, and preference-based requests are outside this definition. The typed graph induces a set $\mathcal{I}(s)$ of method-anchored realizations, each containing one method anchor and its schema-compatible supporting nodes. A capability witness is a set of supporting branches that covers every $r\in\mathcal{R}_q$ within one $I\in\mathcal{I}(s)$; branches assigned to different realizations are never combined.

\paragraph{\textbf{Graph encoding.}}
Relation-aware message passing combines node-text, type, and layer information to encode the candidate graph~\cite{schlichtkrull2018modeling}. A query-conditioned scoring head assigns each node or relation $z$ an evidence strength $m(q,z)\in[0,1]$, computed from its current element representation and sigmoid-normalized before witness aggregation. These scores weight each branch in the Max--Min computation. Ranking feedback calibrates these node representations and relation weights (Section~\ref{sec:graph-calibration}).

\paragraph{\textbf{Graph-aware score.}}
The frozen encoder embeds queries, skills, and nodes. The reranker uses a hidden dimension of 128, two relation-aware layers, and a dropout rate of 0.15. Its score is:
\begin{equation}
    S_G(q,s,G_s)=
    f_{\theta}\!\left(\mathbf q,\mathbf t_s,
    \mathbf p_q(G_s),{\phi}(q,G_s),\mathbf b(q,s)\right).
\end{equation}
Here $\mathbf t_s$ represents skill text; $\mathbf p_q$ concatenates query-attentive, max, and mean graph pooling; ${\phi}$ contains witness, branch, node-match, typed-edge, and attention summaries; and $\mathbf b$ contains first-stage features. The two-layer GELU MLP outputs $S_G$. Graph and retrieval scores are normalized within each pool and fused with $\alpha=0.70$.

\paragraph{\textbf{Witness operationalization.}}
We extract query requirements and match same-type canonical nodes (Appendix~\ref{app:implementation-details}); unmatched requirements remain uncovered. The extractor assigns each alternative procedure an implementation ID, retained by its node and edge occurrences. Canonical nodes share representations while retaining these IDs. Paths stay within one realization and cannot traverse another method anchor. The witness and branch scores enter ${\phi}(q,G_s)$ as auxiliary evidence for skill reranking.

\paragraph{\textbf{Max--Min witness semiring.}}

We follow semiring graph reasoning~\cite{zhu2021neural} and score branches using the Max--Min operations $a\oplus b=\max(a,b)$ and $a\otimes b=\min(a,b)$. The requirement and realization aggregations below enforce joint coverage.
The $\min$ operator limits each supporting branch by its weakest node or relation, while the $\max$ operator selects the strongest alternative branch for the same requirement. A second minimum then enforces conjunction across all requirements covered by one realization, and a final maximum selects among alternative realizations.

\begin{figure}[htbp]
    \centering
    \includegraphics[width=\columnwidth]{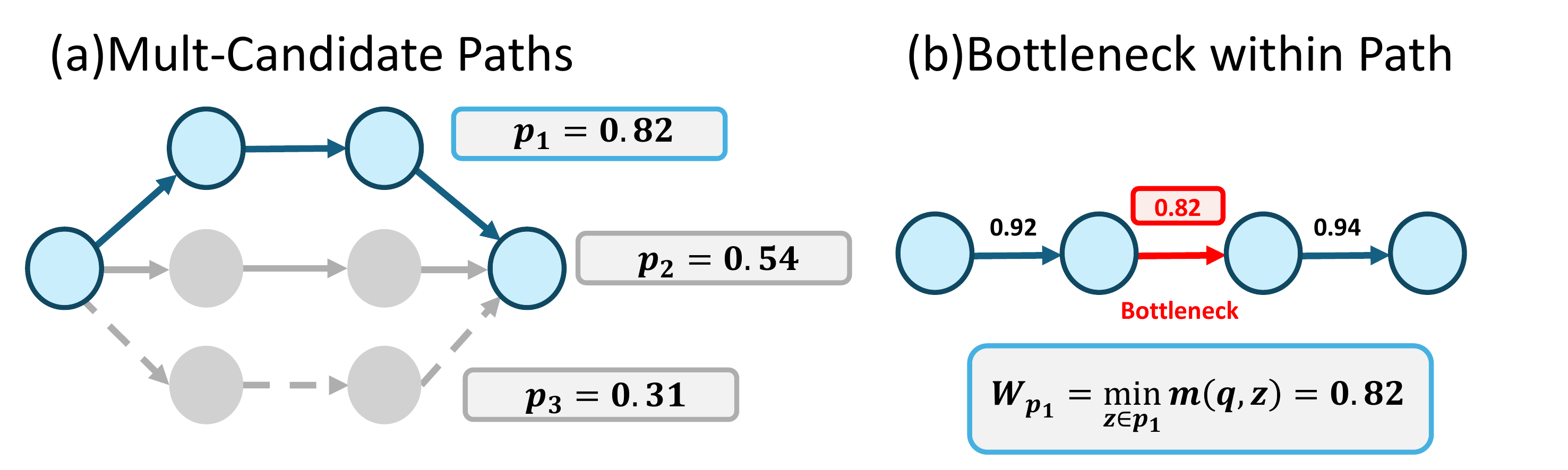}
    \caption{Max--Min path scoring: each branch is limited by its weakest node or relation.}
    \Description{Candidate supporting paths are compared by their bottleneck scores. The strongest path for each requirement contributes to a multi-branch witness anchored to one implementation.}
    \label{fig:semiring-witness}
\end{figure}

For a realization $I$, let $\mathcal{P}_I(r)$ contain the schema-valid paths from its method anchor to nodes that support requirement $r$. The bottleneck score of the strongest supporting branch for $r$ is:
\begin{equation}
    B(q,r,I)=\max_{p\in\mathcal{P}_I(r)}\min_{z\in p}m(q,z),
\end{equation}
where $B(q,r,I)=0$ when no such branch exists. We then require the same realization to cover the complete requirement set and select the strongest valid realization:
\begin{equation}
\begin{aligned}
    W_I(q,s)&=\min_{r\in\mathcal{R}_q} B(q,r,I),\\
    W(q,s)&=\max_{I\in\mathcal{I}(s)} W_I(q,s).
\end{aligned}
\label{eq:conjunctive-witness}
\end{equation}
Thus, a missing requirement makes the realization-level witness zero, strong evidence for one requirement cannot compensate for weak evidence for another, and branches from different realizations cannot be combined.

\paragraph{\textbf{Query-conditioned semiring path closure.}}

For each realization $I$, closure uses its schema-compatible edges $E_I^{\pm}$~\cite{mohri2002semiring} and the path restrictions in Appendix~\ref{app:witness-instantiation}. For nodes $v_i$ and $v_j$ in $I$, define:
\begin{equation}
    [\mathbf{A}_{q,I}]_{ij}
    =\bigoplus_{r:(v_i,r,v_j)\in E_I^{\pm}}
    \left(m(q,v_i)\otimes m(q,(v_i,r,v_j))\otimes m(q,v_j)\right),
\end{equation}
where parallel typed relations are combined by $\oplus$. Matrix
multiplication over the Max--Min semiring is defined as follows:
\begin{equation}
    [\mathbf{A}\otimes\mathbf{B}]_{ij}
    =\bigoplus_k\left(\mathbf{A}_{ik}\otimes\mathbf{B}_{kj}\right)
    =\max_k\min\!\left(\mathbf{A}_{ik},\mathbf{B}_{kj}\right).
\end{equation}
Consequently, the $(i,j)$ entry of $\mathbf{A}_{q,I}^{\otimes \ell}$ gives the
strongest bottleneck value among all length-$\ell$ paths from $v_i$ to
$v_j$. Using $L=4$ yields the truncated closure:
\begin{equation}
    \mathbf{A}_{q,I}^{(\leq L)}
    =\mathbf{A}_{q,I}\oplus\mathbf{A}_{q,I}^{\otimes 2}
    \oplus\cdots\oplus\mathbf{A}_{q,I}^{\otimes L}.
\end{equation}
For illustration only, with unrestricted path lengths, let $D_{ij}^{(0)}=[\mathbf{A}_{q,I}]_{ij}$ and let
$D_{ij}^{(k)}$ denote the best bottleneck value when only
$v_1,\ldots,v_k$ may be used as intermediate nodes. This closure admits
the dynamic-programming recurrence:
\begin{equation}
    D_{ij}^{(k)}
    =\max\!\left(
    D_{ij}^{(k-1)},
    \min\!\left(D_{ik}^{(k-1)},D_{kj}^{(k-1)}\right)
    \right),
\end{equation}
illustrating bottleneck extension and alternative selection. In practice, $B(q,r,I)$ uses the truncated, schema-valid paths above, with a zero-edge branch for a matched method anchor; Equation~\ref{eq:conjunctive-witness} then enforces requirement conjunction and realization selection.

\paragraph{\textbf{Differentiable relaxation.}}

Because strict $\min$ and $\max$ operations yield sparse gradients, we use a log-sum-exp approximation for the minimum within each branch:
\begin{equation}
    \operatorname{SoftMin}_{\tau}(x_1,\ldots,x_n)
    =-\tau\log\left(
    \frac{1}{n}\sum_{i=1}^{n}\exp\left(-\frac{x_i}{\tau}\right)
    \right).
\end{equation}
The remaining extrema are relaxed analogously and approach $\min$ and $\max$ as $\tau\rightarrow0$. Incomplete realizations are hard-masked to zero and excluded from smooth aggregation; if none fully covers the requirements, the skill witness is zero.

\paragraph{\textbf{Capability witness objective.}}

For a positive skill $s^+$ and $M$ hard negatives $\{s_i^-\}_{i=1}^{M}$, define $\Delta_i=W(q,s^+)-W(q,s_i^-)$. We optimize the capability witness loss:
\begin{equation}
    \mathcal{L}_{\mathrm{witness}}
    =\frac{1}{M}\sum_{i=1}^{M}
    \log\left(1+\exp(\gamma_w-\Delta_i)\right),
\end{equation}
where $\gamma_w$ is the witness margin. This objective encourages a positive skill to provide a stronger requirement-covering witness than semantically similar hard negatives.

\subsection{Ranking-Feedback Graph Calibration}
\label{sec:graph-calibration}

Extracted capability graphs may contain incorrect mappings or relations that distort candidate ranking. We use ranking feedback to estimate each element's contribution to candidate discrimination and calibrate its influence. This connects evidence calibration to the skill-selection objective. Earlier graph-learning methods jointly refine structure and representations~\cite{chen2020iterative,jin2020graph}; preference-based tool retrieval learns to reflect the preferences of downstream models~\cite{molfetta-etal-2025-ports}. We distinguish two signals: \emph{structural reliability}, a query-independent estimate of document support for a mapping or relation, and \emph{query relevance}, which measures the element's usefulness for a particular query.

\paragraph{\textbf{Element masking and contribution evaluation.}}

Document support and query relevance alone do not reveal whether an element helps distinguish a positive skill from a semantically similar hard negative. Masking each element reveals its contribution: the change in ranking margin supervises how strongly it should influence each positive--negative candidate comparison.

Given a query $q$, a positive skill $s^+$, and a hard negative $s^-$, let $M(q,s^+,s^-;G)$ denote their margin under the Graph-Aware Score. To estimate the contribution of a graph element $z$, we temporarily mask it and measure the change in the margin:
\begin{equation}
    u_{q,z}=M(q,s^+,s^-;G)-M(q,s^+,s^-;G\setminus z).
\end{equation}
A positive $u_{q,z}$ means that masking $z$ weakens the correct ranking and that $z$ provides useful ranking evidence. A negative value means that removing $z$ improves the margin, indicating that the element may be noisy or misleading. This ranking utility is not itself a label of extraction correctness.

We convert masking utility into a bounded contribution target and incorporate a query-relevance prior to regularize calibration:
\begin{equation}
\mathcal{L}_{\mathrm{attribution}}
        =\frac{1}{|Z_s|}\sum_{z\in Z_s}\operatorname{BCE}(g_{q,z},y_{q,z}).
\end{equation}
Here, $y_{q,z}$ combines normalized masking utility with the query-relevance prior and is treated as a stop-gradient target. Target construction details are provided in Appendix~\ref{app:calibration-target}. Ranking feedback can therefore adjust how much each extracted node or relation contributes without immediately rewriting the discrete graph.

\paragraph{\textbf{Weight calibration.}}

An element may be unhelpful because its extracted structure is unreliable or because it is irrelevant to the current query. Treating these cases identically would suppress valid information that may be useful for other queries. We therefore distinguish document support from query relevance when weighting graph elements.

For each element $z$, a structural MLP estimates confidence $r_z$ from document support. Shared-node features compare canonical and raw representations; edge features encode endpoints and relation type. A query-conditioned MLP estimates relevance $a_{q,z}$. Their product, $g_{q,z}=r_z a_{q,z}$, weights the element's contribution. Low query relevance alone does not imply a structural error. We interpret $r_z$ as a query-independent structural confidence prior and evaluate its empirical association with extraction errors in Section~\ref{app:feedback-robustness}.

\paragraph{\textbf{Graph calibration.}}

A questionable canonical mapping should not erase the original information that distinguishes a skill, and an unreliable relation should not dominate its representation. Learned calibration signals therefore adjust node representations and relation weights within the extracted graph.

For shared nodes, calibration interpolates between the canonical mapping and the original extracted representation:
\begin{equation}
    \mathbf{h}'_{q,v}=(1-\rho_{q,v})\mathbf{h}^{\mathrm{canon}}_v
    +\rho_{q,v}\mathbf{h}^{\mathrm{raw}}_v,
\end{equation}
where $\rho_{q,v}\in[0,1]$ controls the interpolation toward the raw representation, increasing as canonical grounding weakens and query-dependent importance grows. Reliable mappings retain the unified representation; unreliable mappings fall back toward the raw node and preserve local skill information. Edge reliability and query relevance control relation-aware propagation through $g_{q,e}$. The calibrated representations and relation weights determine the updated strengths $m(q,z)$ used in reranking and witness scoring.

\subsection{Training and Inference}

\begin{table*}[tbp]
  \centering
  \setlength{\defaultaddspace}{2.5pt}
  \caption{Retrieval and reranking on SkillRet (percentages). Meta denotes skill metadata. Underlines mark the best published-protocol reranker; matched comparisons appear in a separate group.}
  \label{tab:main-results}
    \begin{tabular*}{\textwidth}{@{\extracolsep{\fill}}llcrrrr@{}}
      \toprule
      Method & Skill input & Model size & NDCG@5 & NDCG@10 & Recall@10 & Comp.@10 \\
      \midrule
      \addlinespace
      \multicolumn{7}{l}{\textit{RAG retrieval baselines}} \\
      Naive-RAG & Full SKILL.md & --- & 56.46 & 59.10 & 67.18 & 52.09 \\
      HippoRAG 2 & Full SKILL.md & --- & 44.80 & 47.15 & 55.55 & 39.23 \\
      RAPTOR & Full SKILL.md & --- & 59.23 & 62.16 & 69.08 & 53.35 \\
      \addlinespace
      \multicolumn{7}{l}{\textit{General-purpose rerankers}} \\
      jina-reranker-v2 & Meta & 0.3B & 69.92 & 73.14 & 83.93 & 73.06 \\
      bge-reranker-v2-m3 & Meta & 0.6B & 53.99 & 59.79 & 77.53 & 62.60 \\
      Qwen3-Reranker-0.6B & Meta & 0.6B & 72.84 & 75.81 & 85.86 & 75.48 \\
      Qwen3-Reranker-4B & Meta & 4.0B & 73.24 & 76.20 & 86.18 & 76.09 \\
      Qwen3-Reranker-8B & Meta & 8.0B & 73.21 & 76.09 & 86.48 & 76.53 \\
      \addlinespace
      \multicolumn{7}{l}{\textit{Skill-specific rerankers}} \\
      SkillRet-Reranker-0.6B & Meta & 0.6B & 80.71 & 82.18 & 87.61 & \underline{78.95} \\
      SkillRouter-Reranker-0.6B & Full SKILL.md & 0.6B & \underline{81.37} & \underline{82.92} & \underline{87.69} & 78.35 \\
      R3-Reranker-0.6B & Full SKILL.md & 0.6B & 80.44 & 81.93 & 87.15 & 78.39 \\
      \addlinespace
      \multicolumn{7}{l}{\textit{Graph-based related work}} \\
      SkillDAG & Full SKILL.md & --- & 75.59 & 77.19 & 83.39 & 71.72 \\
      \addlinespace
      \multicolumn{7}{l}{\textit{Matched internal baselines}} \\
      SkillRet-Embedding-0.6B (baseline) & Meta & 0.6B & 75.57 & 78.03 & 85.42 & 75.09 \\
      Text-only Pairwise Reranker & Meta & --- & 80.98 & 84.07 & 93.39 & 87.41 \\
      Full SKILL.md Reranker & Full SKILL.md & --- & 79.75 & 83.06 & 93.24 & 87.22 \\
      Graph-as-Text Reranker & Meta + Graph Text & --- & 81.10 & 84.09 & 93.17 & 86.66 \\
      \addlinespace[2pt]
      \textbf{SKALOS} & Meta + Graph & 2.0M + frozen 0.6B & \textbf{84.33} & \textbf{86.76} & \textbf{94.57} & \textbf{89.59} \\
      \bottomrule
    \end{tabular*}
\end{table*}

Let $\Delta_i^S=S_G(q,s^+,G^+)-S_G(q,s_i^-,G_i^-)$ denote the Graph-Aware Score difference between a positive skill and the $i$-th hard negative. The pairwise ranking loss is defined as follows:
\begin{equation}
    \mathcal{L}_{\mathrm{rank}}
    =\frac{1}{M}\sum_{i=1}^{M}
    \log\left(1+\exp(-\Delta_i^S)\right).
\end{equation}
We train the graph reranker with the combined objective:
\begin{equation}
    \mathcal{L}_{\mathrm{graph}}=\mathcal{L}_{\mathrm{rank}}
    +\lambda_w\mathcal{L}_{\mathrm{witness}}.
\end{equation}
We then freeze its parameters and train a lightweight graph-calibration module using the masking-derived attribution objective. At inference, calibration is applied to the candidate graph before the frozen reranker and witness scorer are evaluated, keeping calibration and witness scoring active throughout inference.

Initialization uses 63,259 training queries, followed by graph-native updates on 1,000 queries and 8,120 pairs. Negatives come exclusively from the top-20 pool: half are the highest-ranked non-relevant candidates, and the rest are sampled from the remaining non-relevant candidates. Calibration collects 11,914 masking examples from 1,000 training queries, keeping the reranker frozen.

At inference time, the learned Graph-Aware Score is combined with the first-stage retrieval score using a fixed coefficient $\alpha$:
\begin{equation}
    S_{\mathrm{final}}(q,s)
    =(1-\alpha)\widetilde{S}_{\mathrm{ret}}(q,s)
    +\alpha\widetilde{S}_G(q,s,G_s).
\end{equation}
Both scores are normalized within each candidate pool. The witness supplies an auxiliary signal of joint requirement coverage within one implementation.

\section{Experiments}
We evaluate SKALOS through six research questions:
\begin{itemize}
    \item \textbf{RQ1:} How well does SKALOS rerank skills?
    \item \textbf{RQ2:} Do gains transfer across rerankers?
    \item \textbf{RQ3:} Which skill representation yields the best reranking performance?
    \item \textbf{RQ4:} How much does each component contribute?
    \item \textbf{RQ5:} Can feedback detect and correct graph errors?
    \item \textbf{RQ6:} Do witnesses improve agent task success?
\end{itemize}

\subsection{Experimental Setup}

The frozen SkillRet-Embedding-0.6B retriever~\cite{cho2026skillret} supplies the top 20 candidate skills for reranking. SKALOS has about 2.0M trainable parameters: 1.999M for reranking and 1,089 for calibration.
The text encoder remains frozen. A frozen Qwen2.5-7B-Instruct model extracts skill graphs offline and requirements from each query.
The reranker is initialized on the full official training split, then fine-tuned as detailed in Appendix~\ref{app:implementation-details}. We fix the fusion weight at 0.70 using training data. Evaluation metrics are reported as percentages.

\paragraph{\textbf{Baselines.}}
We compare SKALOS with a broad set of representative baselines, including SkillRet-Embedding-0.6B and SkillRet-Reranker-0.6B~\cite{cho2026skillret}; general-purpose rerankers (Jina~\cite{jina2024rerankerv2}, BGE~\cite{li2024llama2vec,chen2024bge}, and Qwen3~\cite{zhang2025qwen3}); skill-specific rerankers (SkillRouter~\cite{zheng2026skillrouterskillroutingllm} and R3~\cite{wang2026skill}); the graph-based SkillDAG~\cite{bai2026skilldag}; and full-corpus RAG methods (Naive-RAG, HippoRAG~2~\cite{pmlr-v267-gutierrez25a}, and RAPTOR~\cite{sarthi2024raptor}). Matched baselines represent each skill using metadata, its full document, serialized capability nodes, or a serialized graph with the same frozen encoder, top-20 pools, and training protocol.

\paragraph{\textbf{Datasets.}}

We evaluate on SkillRet~\cite{cho2026skillret}, a skill-retrieval benchmark with 63,259 training queries, 4,997 test queries, and binary relevance labels. For downstream generalization, we additionally use ALFWorld~\cite{shridhar2020alfworld} household tasks.

\paragraph{\textbf{Metrics.}}

We follow earlier evaluations~\cite{cho2026skillret,zheng2026skillrouterskillroutingllm,wang2026skill,bai2026skilldag} and report \textbf{NDCG@5/10} for ranking quality~\cite{jarvelin2002cumulated}, \textbf{Recall@10}, and \textbf{Completeness@10}. These metrics evaluate ranking quality and the coverage of relevant skills. All metrics are averaged across evaluation queries; detailed metric definitions and significance tests are reported in Appendix~\ref{app:evaluation-protocol}.

\subsection{Main Results}

To answer RQ1, Table~\ref{tab:main-results} reports overall results. On the matched candidate pool, SKALOS outperforms the text-only, full-document, and graph-as-text rerankers on all four metrics; published and full-corpus RAG scores are reference-only because their protocols differ. Using full documents does not improve performance over using metadata alone, and serialized graph text yields little gain. SKALOS instead tests how capability evidence fits together.

All three matched comparisons yield positive paired-bootstrap 95\% confidence intervals for NDCG@10 gains, with Holm-adjusted $p<0.001$ (Appendix~\ref{app:main-uncertainty}).

\begin{figure}[htbp]
  \centering
  \includegraphics[width=\columnwidth]{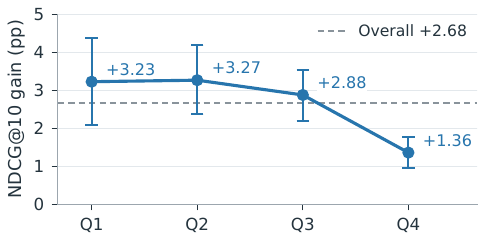}
  \caption{NDCG@10 gains by score-gap quartile (Q1: smallest gap). Error bars show 95\% CIs.}
  \Description{Graph reranking improves NDCG at 10 in every score-gap quartile. Gains are 3.23, 3.27, 2.88, and 1.36 percentage points from the smallest to largest gap. Error bars show paired bootstrap confidence intervals.}
  \label{fig:natural-ambiguity}
\end{figure}

\paragraph{\textbf{First-stage ambiguity.}}
We group all 4,997 test queries by the gap between their top two first-stage scores (Figure~\ref{fig:natural-ambiguity}). SKALOS gains 3.23 and 3.27 NDCG@10 points in Q1 and Q2, respectively, versus 1.36 in Q4. Smaller gaps indicate first-stage ambiguity, not capability insufficiency; baseline difficulty also varies. Intervals are estimated from 5,000 paired bootstraps over the full set of test queries.

\subsection{Cross-Reranker Generalization}

For RQ2, five rerankers span different architectures and model sizes, each with a separately trained graph module. On common top-20 pools, Base uses re-evaluated reranker scores; +Embed and +Graph fuse these with first-stage embedding scores and the corresponding Graph-Aware Score, respectively.

\begin{figure}[htbp]
    \centering
    \includegraphics[width=\columnwidth]{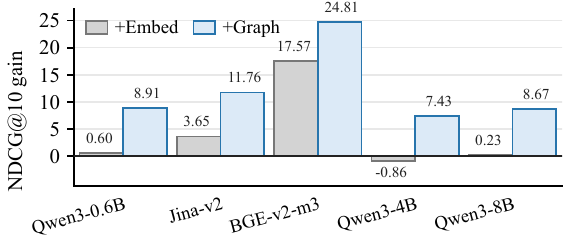}
    \caption{NDCG@10 gains from embedding fusion (+Embed) and graph augmentation (+Graph) across base rerankers.}
    \Description{Grouped bars compare embedding-score fusion and Graph-Aware Score fusion across five rerankers. Each graph module is trained separately for its reranker. Graph augmentation improves all five rerankers and consistently exceeds embedding-score fusion.}
    \label{fig:cross-reranker-gain}
\end{figure}

Figure~\ref{fig:cross-reranker-gain} shows that +Graph improves NDCG@10 by 7.43--24.81 points over Base and by 7.24--8.45 points over +Embed. These consistent gains indicate that embedding-score fusion alone does not explain the improvement. Full results appear in Appendix~\ref{app:cross-reranker}.

\subsection{Skill Representation Comparison}

To answer RQ3, we compare metadata, full-document, serialized-node, and serialized-graph inputs against explicit graph modeling.

\begin{table}[htbp]
  \centering
  \small
  \caption{Comparison of skill representations.}
  \label{tab:representation-ablation}
  \setlength{\tabcolsep}{3pt}
    \begin{tabular*}{\columnwidth}{@{\extracolsep{\fill}}lrrr@{}}
      \toprule
      Method & NDCG@10 & Recall@10 & Comp.@10 \\
      \midrule
      First-stage Retrieval & 78.03 & 85.42 & 75.09 \\
      Text-only Pairwise Reranker & 84.07 & 93.39 & 87.41 \\
      Full SKILL.md Reranker & 83.06 & 93.24 & 87.22 \\
      Node-as-Text Reranker & \underline{84.21} & \underline{93.48} & \underline{87.43} \\
      Graph-as-Text Reranker & 84.09 & 93.17 & 86.66 \\
      \textbf{SKALOS} & \textbf{86.76} & \textbf{94.57} & \textbf{89.59} \\
      \bottomrule
    \end{tabular*}
\end{table}

SKALOS exceeds Node-as-Text, the strongest alternative in
Table~\ref{tab:representation-ablation}, by 2.55 NDCG@10 points.

We directly test requirement coverage with 93 source-disjoint skills. Each four-candidate pool has one complete implementation and three similar negatives: one omits an artifact, one omits a constraint, and one splits evidence across implementations. We apply 10\% node and relation dropout.

\begin{table}[htbp]
  \centering
  \small
  \caption{Requirement-coverage stress test after removing 10\% of graph nodes and relations.}
  \label{tab:witness-coverage}
  \setlength{\tabcolsep}{3pt}
  \begin{tabular*}{\columnwidth}{@{\extracolsep{\fill}}lrrr@{}}
    \toprule
    Method & NDCG@4 & Pair Acc. & Margin \\
    \midrule
    Text Reranker & 94.97 & 95.40 & 0.0129 \\
    \textbf{Witness} & \textbf{97.27} & \textbf{97.53} & \textbf{0.0391} \\
    \bottomrule
  \end{tabular*}
\end{table}

The witness gains 2.29 NDCG@4 points, with a 95\% CI of $[1.54,3.09]$ from source-clustered bootstraps. It also widens the positive--negative margin (Table~\ref{tab:witness-coverage}). This controlled test checks joint support within a single procedure and complements the relevance-based SkillRet evaluation.

\subsection{Component Ablation Studies}

To answer RQ4, we retrain each ablation with the same data and supervision. W/O Graph retains query, skill-text, and first-stage features but removes graph pooling, summaries, and witnesses; other variants disable only the named channel.

\begin{table}[htbp]
  \centering
  \small
  \caption{Component ablation results.}
  \label{tab:component-ablation}
  \setlength{\tabcolsep}{3pt}
    \begin{tabular*}{\columnwidth}{@{\extracolsep{\fill}}lrrr@{}}
      \toprule
      Setting & NDCG@10 & Recall@10 & Comp.@10 \\
      \midrule
      \textbf{SKALOS} & \textbf{86.76} & \textbf{94.57} & \textbf{89.59} \\
      W/O Graph & 83.32 & 92.67 & 86.48 \\
      W/O Edge & 80.35 & 93.00 & 86.58 \\
      W/O Relation Type & 81.23 & \underline{93.34} & \underline{87.26} \\
      W/O Node Type & 81.77 & 92.96 & 86.62 \\
      W/O Hierarchy & 80.83 & 93.12 & 86.96 \\
      W/O Attention & 79.26 & 92.03 & 84.86 \\
      W/O Witness & 84.07 & 90.70 & 82.95 \\
      W/O Graph Calib. & \underline{84.28} & 88.29 & 78.48 \\
      W/O Text & 26.38 & 50.35 & 33.93 \\
      \bottomrule
    \end{tabular*}
\end{table}

Every removal hurts performance (Table~\ref{tab:component-ablation}). Removing graph-derived signals lowers NDCG@10 by 3.44 points; edge, type, and witness ablations further support relational and conjunctive modeling. Removing calibration from the original extracted graphs lowers NDCG@10 by 2.48 points.

Removing the witness also reduces Recall@10 from 94.57 to 90.70 and Completeness@10 from 89.59 to 82.95. Its effect therefore extends beyond ordering quality to retrieval coverage in the fixed pool. These labels identify relevant skills without certifying executable support; the separate coverage stress test provides the controlled check of implementation consistency.

\subsection{Ranking-Feedback Graph Calibration}

To answer RQ5, we introduce test-time extraction errors by replacing a query-relevant shared-node label with a plausible same-type label from its cluster. We measure detection by AUPRC, error Recall@20\%, and Enrichment@20\% ($1\times$ is random selection); definitions appear in Appendix~\ref{app:feedback-robustness}.

\begin{table}[htbp]
  \centering
  \small
  \caption{Error detection under evidence corruption.}
  \label{tab:graph-feedback-diagnostics}
  \setlength{\tabcolsep}{3pt}
    \begin{tabular*}{\columnwidth}{@{\extracolsep{\fill}}lrrr@{}}
      \toprule
      Corruption & AUPRC & Recall@20\% & Enrichment@20\% \\
      \midrule
      25\% & $0.799\pm0.015$ & $78.3\%\pm1.9\%$ & $(3.92\pm0.09)\times$ \\
      50\% & $0.593\pm0.004$ & $57.6\%\pm0.7\%$ & $(2.88\pm0.04)\times$ \\
      75\% & $0.369\pm0.014$ & $37.9\%\pm1.8\%$ & $(1.89\pm0.09)\times$ \\
      Random & 0.090 & 20.0\% & $1.00\times$ \\
      \bottomrule
    \end{tabular*}
\end{table}

Table~\ref{tab:graph-feedback-diagnostics} shows that the lowest-confidence nodes are enriched 3.92$\times$ and 2.88$\times$ for inserted errors at 25\% and 50\% corruption; detection weakens but remains above random at 75\%.

We also introduce errors identified by human inspection of extracted graphs. ``Budget'' is the targeted error fraction and ``Perturbed'' the realized fraction. Pair and score recovery measure reductions in deviations from clean ranking margins and candidate scores. W/O Attribution Loss replaces masking utility with the relevance prior; W/O Graph Calibration bypasses correction.

\begin{table}[htbp]
  \centering
  \small
  \caption{Calibration under human-annotated errors. Error budgets and recovery scores are percentages.}
  \label{tab:feedback-mechanism}
  \setlength{\tabcolsep}{3pt}
    \begin{tabular*}{\columnwidth}{@{\extracolsep{\fill}}lrrrr@{}}
      \toprule
      Setting & \shortstack{Perturbed\\(\%)} & \shortstack{Enrich.\\@20\% ($\times$)} & \shortstack{Pair rec.\\(\%)} & \shortstack{Score rec.\\(\%)} \\
      \midrule
      Full (10\%) & 9.89 & 3.078 & 34.38 & 32.42 \\
      Full (20\%) & 20.10 & 3.029 & 33.58 & 32.45 \\
      Full (30\%) & 30.45 & 2.944 & 35.38 & 32.01 \\
      \midrule
      W/O Attr. (20\%) & 20.10 & --- & 29.60 & 29.33 \\
      \bottomrule
    \end{tabular*}
\end{table}

Table~\ref{tab:feedback-mechanism} shows that error enrichment remains near 3$\times$ random selection across budgets, while calibration recovers 33.58--35.38\% of pair-margin and 32.01--32.45\% of score deviations. Removing attribution loss reduces recovery by 3.98 and 3.12 percentage points at the 20\% budget, confirming the additional value of ranking feedback beyond query relevance. Without graph calibration, these deviations remain uncorrected.

\paragraph{\textbf{Scoring cost.}}
With cached text embeddings and graph inputs, SKALOS scores a top-20 pool in 134.85\,ms, or 146.97\,ms with edge augmentation. Peak GPU memory, including the graph cache, is 181.64 and 184.16\,MiB, respectively. Edge augmentation introduces only 12.12\,ms additional latency and 2.52\,MiB memory overhead, indicating modest computational overhead for richer structural evidence. These timings cover scoring alone and exclude extraction (Appendix~\ref{app:efficiency}).

\subsection{Agent Execution Generalization}

To answer RQ6, each ALFWorld task pairs a valid skill with three textually similar negatives that corrupt one critical requirement. The selected skill guides the same Qwen2.5-7B controller; success is ALFWorld's completion signal.

\begin{figure}[htbp]
  \centering
  \includegraphics[width=\columnwidth]{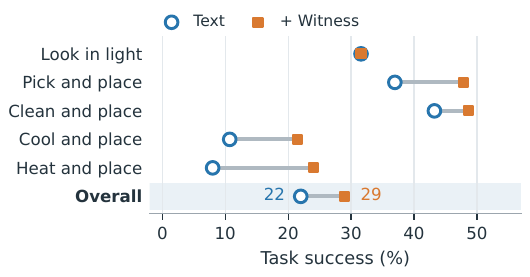}
  \caption{ALFWorld task success by task family.}
  \Description{Task-family success counts compare text relevance with capability witness selection on 200 ALFWorld tasks. Total successes rise from 44 to 58 with the same controller and candidate pools.}
  \label{fig:alfworld-success}
\end{figure}

Overall, witness selection succeeds on 58/200 tasks (29\%) versus 44/200 (22\%) for text relevance. The paired outcomes include 18 witness-only and four text-only successes, giving an exact two-sided McNemar test $p=0.004$ (Appendix~\ref{app:alfworld-execution}).

\section{Conclusion}

SKALOS reranks skills using explicit capability evidence. Its graph aligns heterogeneous descriptions while retaining skill-specific details; witness learning checks joint requirement support within one implementation; and calibration limits the influence of unreliable evidence. SKALOS gains 2.68 NDCG@10 points over the matched text-only reranker and raises controlled ALFWorld success from 22\% to 29\%. Gains across five rerankers and the component ablations
support the utility of the proposed framework.
Controlled perturbations further show that ranking feedback helps recover scores and margins distorted by extraction errors. These results suggest that requirement coverage, implementation consistency, and evidence reliability provide complementary signals for distinguishing semantically similar skills. Future work will extend witness modeling to optional and disjunctive requirements and evaluate these extensions with more diverse skills and broader agent execution settings.

\bibliographystyle{ACM-Reference-Format}
\bibliography{references}

\appendix
\section{Reproducibility}
\label{app:reproducibility}

\subsection{Evaluation Protocol and Ranking Metrics}
\label{app:evaluation-protocol}

Controlled reranking uses the frozen SkillRet-Embedding-0.6B encoder and its top-20 candidate pool on all 4,997 test queries. Public baselines retain published protocols; RAG methods retrieve from the full corpus and are not matched comparisons. Feedback diagnostics use separate 300-query evaluations.

\paragraph{\textbf{Ranking metrics.}}
Using binary relevance labels, NDCG@$k$ discounts gain at rank $i$ by $1/\log_2(i+1)$ and normalizes it by ideal gain~\cite{jarvelin2002cumulated}. Recall@$k$ is the fraction of labeled relevant skills retrieved in the top $k$. Completeness@$k$ equals one only when the top $k$ contains every labeled relevant skill~\cite{zhang2024data}. It measures retrieval coverage without certifying capability sufficiency. We report query-level averages as percentages.

\subsection{Model and Optimization Details}
\label{app:implementation-details}

\paragraph{\textbf{Graph construction.}}
The offline Qwen2.5-7B-Instruct extractor uses deterministic decoding and validated JSON, extracting four node types before predicting typed edges with exact node endpoints; invalid output receives one format-repair pass. Within each cluster, the frozen encoder embeds raw phrases. Phrases of the same type merge at cosine similarity $\geq0.86$. Query requirements use the same frozen typed extractor and matching configuration; unmatched requirements remain uncovered. Nodes supported by at least two skills enter the shared layer; single-skill nodes remain local to preserve evidence specific to each candidate.

\begin{table}[htbp]
  \centering
  \footnotesize
  \caption{Capability-graph cluster and node statistics.}
  \label{tab:cluster-graph-statistics}
  \setlength{\tabcolsep}{2.7pt}
  \begin{tabular*}{\columnwidth}{@{\extracolsep{\fill}}lrrrrrr}
    \toprule
    Split & Clusters & Skills/cluster & Cap. & Method & Art. & Const. \\
    \midrule
    Train & 100 & 101.23 & 420.87 & 19.25 & 17.95 & 20.03 \\
    Test  & 95  & 70.11  & 298.88 & 14.58 & 13.15 & 14.40 \\
    \bottomrule
  \end{tabular*}
\end{table}

\paragraph{\textbf{Optimization.}}
All graph-reranker parameters remain trainable during initialization and graph-native updates. Negatives are drawn exclusively from the top-20 pool: half are the highest-ranked non-relevant candidates, and the rest are sampled from the remaining non-relevant candidates.

Table~\ref{tab:training-hyperparameters} summarizes these settings.

\begin{table}[htbp]
  \centering
  \footnotesize
  \caption{Hyperparameters for graph-reranker training.}
  \label{tab:training-hyperparameters}
  \setlength{\tabcolsep}{3.5pt}
  \begin{tabular*}{\columnwidth}{@{\extracolsep{\fill}}p{0.23\columnwidth}p{0.43\columnwidth}p{0.28\columnwidth}@{}}
    \toprule
    Component & Hyperparameter & Value \\
    \midrule
    Text encoder & Status & Frozen \\
    Graph encoding & Hidden size / layers / dropout & 128 / 2 / 0.15 \\
    Optimization & Optimizer / weight decay & AdamW / $10^{-4}$ \\
                 & Gradient clip / accumulation & 1.0 / 8 \\
    Initialization & Epochs / LR / max negatives & 3 / $5\!\times\!10^{-4}$ / 8 \\
    Graph training & Epochs / LR / max negatives & 2 / $10^{-4}$ / 4 \\
                   & Queries / training pairs & 1,000 / 8,120 \\
    Ranking / witness & $\gamma_r$ / $\lambda_w$ / $\gamma_w$ / $\tau$ & 0 / 0.05 / 0.05 / 0.10 \\
    Calibration & Queries / masking examples & 1,000 / 11,914 \\
                & LR / utility--prior mix / blend & $3\!\times\!10^{-3}$ / 0.8--0.2 / 0.70 \\
                & Utility scale $Q_{0.75}(|u|)$ & 4.145 \\
    Score fusion & $\alpha$ & 0.70 \\
    Reproducibility & Random seed & 42 \\
    \bottomrule
  \end{tabular*}
\end{table}

\paragraph{\textbf{Graph-calibration module.}}
\label{app:calibration-target}
We freeze the trained reranker and collect 11,914 masking examples from 1,000 training queries. The utility scale derived from these examples is $\tau_a=Q_{0.75}(|u|)=4.145$. The 1,089-parameter calibrator uses AdamW at $3\times10^{-3}$ for at most 200 epochs (patience 20). Let $s_v^{\mathrm{str}}$ and $s_{q,v}^{\mathrm{rel}}$ denote the pre-sigmoid outputs of the structural-confidence and query-relevance networks, so $r_v=\sigma(s_v^{\mathrm{str}})$ and $a_{q,v}=\sigma(s_{q,v}^{\mathrm{rel}})$. Before frozen reranking and witness scoring, shared-node representations are updated with $\eta=0.70$:
\begin{equation}
\begin{aligned}
\rho_{q,v}
&=\frac{\eta}{2}\sigma(-s_v^{\mathrm{str}})
  \left[1+\sigma(s_{q,v}^{\mathrm{rel}})\right],\\
\mathbf h'_{q,v}
&=\mathbf h_v^{\mathrm{canon}}
  +\rho_{q,v}\left(\mathbf h_v^{\mathrm{raw}}
  -\mathbf h_v^{\mathrm{canon}}\right).
\end{aligned}
\end{equation}
We construct the attribution target as:
\begin{equation}
\begin{aligned}
    \hat{u}_{q,z}
        &=\frac{1+\operatorname{clip}(u_{q,z}/\tau_a,-1,1)}{2},\\
    y_{q,z}&=0.8\hat{u}_{q,z}+0.2\bar a_{q,z}.
\end{aligned}
\end{equation}

The node-relevance prior uses frozen encoder embeddings:
\begin{equation}
    \bar a_{q,v}=\frac{1+\cos(\mathbf q,\mathbf h_v^{\mathrm{canon}})}{2}.
\end{equation}
Target weights are fixed; gradients do not pass through the target.

\subsection{Witness Instantiation and Scope}
\label{app:witness-instantiation}

Implementation IDs are retained on node and edge occurrences; canonical sharing preserves their membership. Each realization has a single method anchor; paths stay within that realization.

\paragraph{\textbf{Witness computation.}}
Paths stay within one implementation and use at most four declared or inverse typed edges. Allowed relations include \textsc{implements}, \textsc{produces}, and \textsc{constrains}; a matched method anchor permits a zero-edge branch. The complete relation schema and traversal checks are provided in the accompanying code. Incomplete realizations are excluded from smooth aggregation. If all realizations are incomplete, the witness is zero.

\subsection{Statistical Protocol}
\label{app:statistical-protocol}

\paragraph{\textbf{Confidence intervals.}}
With saved per-query predictions, we draw 5,000 same-size query bootstraps (seed 20260908). The 2.5th and 97.5th percentiles of their mean scores give the 95\% CI. Paired gains resample identical IDs for both systems and are absolute percentage-point differences.

\paragraph{\textbf{Significance tests.}}
We apply two-sided Wilcoxon signed-rank tests to per-query differences, using Pratt zero handling and family-wise Holm correction, with NDCG@10 as the primary metric.

\paragraph{\textbf{Run variation.}}
Mean $\pm$ SD summarizes run variation using sample denominator $R-1$; bootstrap CIs describe query uncertainty for a fixed model. Two main-model seeds are reported separately, without seed-level CIs. Table~\ref{tab:seed-stability} lists both runs.
\label{app:seed-stability-compact}

\begin{table}[H]
  \centering
  \footnotesize
  \caption{SKALOS results for two training seeds.}
  \label{tab:seed-stability}
  \setlength{\tabcolsep}{3pt}
  \begin{tabular*}{\columnwidth}{@{\extracolsep{\fill}}lrrr@{}}
    \toprule
    Metric & Seed 42 & Seed 43 & Mean $\pm$ SD \\
    \midrule
    NDCG@5 & 84.331 & 84.560 & $84.446\pm0.162$ \\
    NDCG@10 & 86.758 & 87.032 & $86.895\pm0.194$ \\
    Recall@10 & 94.568 & 94.959 & $94.764\pm0.276$ \\
    Comp.@10 & 89.591 & 90.353 & $89.972\pm0.539$ \\
    \bottomrule
  \end{tabular*}
\end{table}

\subsection{Implementation and Online Efficiency}
\label{app:efficiency}

Table~\ref{tab:online-efficiency} times 300 queries after 20 warm-ups with per-query GPU synchronization on a V100-SXM2 (32\,GB), Xeon Gold 6148, and 128\,GiB RAM, using PyTorch 2.4.0/CUDA 12.1. It compares base and edge-augmented graphs.

\paragraph{\textbf{Efficiency metrics.}}
We measure reranker scoring latency per candidate pool using precomputed text embeddings and graph inputs. P95 denotes the 95th percentile of per-query scoring latency, and QPS measures scoring throughput. GPU memory reports peak allocation of the scoring workload, including the graph cache.

\begin{table}[htbp]
  \centering
  \footnotesize
  \caption{Witness scoring efficiency for two graph inputs, using precomputed text embeddings.}
  \label{tab:online-efficiency}
  \setlength{\tabcolsep}{2.5pt}
    \begin{tabular*}{\columnwidth}{@{\extracolsep{\fill}}lrrrr}
      \toprule
      Graph & Mean (ms) & P95 (ms) & QPS & Memory (MiB) \\
      \midrule
      Base      & 134.85 & 155.60 & 7.42 & 181.64 \\
      Augmented & 146.97 & 168.47 & 6.80 & 184.16 \\
      \bottomrule
    \end{tabular*}%
\end{table}

\paragraph{\textbf{Full-evaluation records.}}
Table~\ref{tab:full-eval-latency} reports historical scoring logs for 4,986 queries with precomputed embeddings, separately from the current 4,997-query evaluation.

\begin{table}[htbp]
  \centering
  \footnotesize
  \caption{Graph-scoring latency from full evaluation with precomputed embeddings.}
  \label{tab:full-eval-latency}
  \setlength{\tabcolsep}{3pt}
  \begin{tabular*}{\columnwidth}{@{\extracolsep{\fill}}lrrr@{}}
    \toprule
    Model/run & Time (s) & ms/query & QPS \\
    \midrule
    Base reranker & 446 & 89.45 & 11.18 \\
    Witness, run 1 & 623 & 124.95 & 8.00 \\
    Witness, run 2 & 674 & 135.18 & 7.40 \\
    \bottomrule
  \end{tabular*}
  \vspace{2pt}
  \parbox{\columnwidth}{\footnotesize Latency and throughput are computed from logged elapsed times. Witness runs are separate evaluation records, not independent training seeds.}
\end{table}

\subsection{Training and Offline Construction Costs}

\paragraph{\textbf{Offline graph construction.}}
Table~\ref{tab:offline-construction-cost} reports clustering and graph-extraction costs for both corpora.

\begin{table}[htbp]
  \centering
  \footnotesize
  \caption{Offline capability-graph construction cost.}
  \label{tab:offline-construction-cost}
  \setlength{\tabcolsep}{2.5pt}
    \begin{tabular*}{\columnwidth}{@{\extracolsep{\fill}}llrrr}
      \toprule
      Split & Stage & Total & Median/Skill & P95/Skill \\
      \midrule
      Train & Clustering & 128.96 min & -- & -- \\
      Train & Graph extraction & 23.21 h & 24.34 s & 38.47 s \\
      Test  & Clustering & 64.63 min & -- & -- \\
      Test  & Graph extraction & 24.25 h & 22.80 s & 28.00 s \\
      \bottomrule
    \end{tabular*}%
\end{table}

\paragraph{\textbf{Model and training costs.}}
Table~\ref{tab:training-cost} separates initialization on all 63,259 queries from the 1,000-query graph update and subsequent calibration; parameter counts exclude the frozen encoder.

\begin{table}[htbp]
  \centering
  \footnotesize
  \caption{Trainable model size and recorded training cost.}
  \label{tab:training-cost}
  \setlength{\tabcolsep}{2.5pt}
    \begin{tabular*}{\columnwidth}{@{\extracolsep{\fill}}lrrl}
      \toprule
      Component & Parameters & Checkpoint & Training record \\
      \midrule
      Base graph reranker & 0.916M & 3.51 MiB & 33.68 h (3 epochs) \\
      Graph-native reranker & 1.999M & 7.65 MiB & 6.07 min (2 epochs) \\
      Graph-calibration module & 1,089 & 9.70 KiB & 65.7 min (200 epochs) \\
      \bottomrule
    \end{tabular*}%
\end{table}

\section{Robustness and Additional Analyses}
\label{app:robustness}

\subsection{Main-Result Uncertainty}
\label{app:main-uncertainty}

Paired NDCG@10 gains over matched baselines range from 2.672 to 3.703 points, with positive CI lower bounds and Holm-adjusted $p<0.001$ (Table~\ref{tab:main-paired-significance}).

\begin{table}[htbp]
  \centering
  \footnotesize
  \caption{Paired NDCG@10 gains of SKALOS over matched internal baselines on all 4,997 test queries.}
  \label{tab:main-paired-significance}
  \setlength{\tabcolsep}{3pt}
  \begin{tabular*}{\columnwidth}{@{\extracolsep{\fill}}lrrr@{}}
    \toprule
    Comparator & Gain & 95\% CI & Holm $p$ \\
    \midrule
    Text-only & 2.684 & [2.263, 3.103] & $<0.001$ \\
    Full SKILL.md & 3.703 & [3.257, 4.174] & $<0.001$ \\
    Graph-as-Text & 2.672 & [2.237, 3.095] & $<0.001$ \\
    \bottomrule
  \end{tabular*}
\end{table}

\subsection{Cross-Reranker Generalization}
\label{app:cross-reranker}

In Table~\ref{tab:cross-reranker}, Base denotes each reranker's scores re-evaluated on the common top-20 pools. +Embed fuses them with embedding retrieval scores; +Graph fuses them with a Graph-Aware Score trained separately for that reranker on SkillRet. These Base scores differ from the published-protocol results in Table~\ref{tab:main-results}.

\begin{table}[htbp]
  \centering
  \footnotesize
  \caption{Graph augmentation results (NDCG@10).}
  \label{tab:cross-reranker}
  \setlength{\tabcolsep}{3pt}
  \begin{tabular*}{\columnwidth}{@{\extracolsep{\fill}}lrrr}
      \toprule
      Reranker & Base & +Graph & $\Delta$ \\
      \midrule
      Qwen3-0.6B & 79.06 & 87.97 & +8.91 \\
      Jina-v2 & 75.91 & 87.67 & \underline{+11.76} \\
      BGE-v2-m3 & 59.79 & 84.60 & \textbf{+24.81} \\
      Qwen3-4B & \textbf{81.10} & \underline{88.53} & +7.43 \\
      Qwen3-8B & \underline{80.11} & \textbf{88.78} & +8.67 \\
      \bottomrule
  \end{tabular*}

\end{table}

All ten gains have positive 95\% CI lower bounds and Holm-adjusted $p<0.001$.

\begin{table}[htbp]
  \centering
  \footnotesize
  \caption{Paired cross-reranker gains on NDCG@10.}
  \label{tab:cross-reranker-confidence}
  \setlength{\tabcolsep}{2.5pt}
  \begin{tabular*}{\columnwidth}{@{\extracolsep{\fill}}llrrr@{}}
    \toprule
    Model & Contrast & Gain & 95\% CI & Holm $p$ \\
    \midrule
    Qwen3-0.6B & vs. Base   & 8.913  & [8.223, 9.599]   & $<.001$ \\
               & vs. +Embed & 8.312  & [7.584, 9.046]   & $<.001$ \\
    Jina-v2    & vs. Base   & 11.758 & [11.033, 12.489] & $<.001$ \\
               & vs. +Embed & 8.106  & [7.369, 8.862]   & $<.001$ \\
    BGE-v2-m3  & vs. Base   & 24.805 & [23.973, 25.660] & $<.001$ \\
               & vs. +Embed & 7.236  & [6.473, 7.995]   & $<.001$ \\
    Qwen3-4B   & vs. Base   & 7.427  & [6.742, 8.091]   & $<.001$ \\
               & vs. +Embed & 8.289  & [7.546, 9.036]   & $<.001$ \\
    Qwen3-8B   & vs. Base   & 8.671  & [7.995, 9.333]   & $<.001$ \\
               & vs. +Embed & 8.446  & [7.696, 9.160]   & $<.001$ \\
    \bottomrule
  \end{tabular*}
\end{table}

\subsection{Controlled Agent-Execution Protocol}
\label{app:alfworld-execution}

We use all 200 distinct ALFWorld \texttt{valid\_train} task directories, none used for selector training; Table~\ref{tab:alfworld-breakdown} gives task-family counts.

\begin{table}[H]
    \centering
    \caption{ALFWorld execution success by task family. Text and + Witness report successful task counts.}
    \label{tab:alfworld-breakdown}
    \footnotesize
    \setlength{\tabcolsep}{3pt}
    \begin{tabular*}{\columnwidth}{@{\extracolsep{\fill}}lrrr@{}}
        \toprule
        Task family & Tasks & Text & + Witness \\
        \midrule
        Look in light & 19 & 6 & 6 \\
        Pick and place & 46 & 17 & 22 \\
        Clean and place & 37 & 16 & 18 \\
        Cool and place & 28 & 3 & 6 \\
        Heat and place & 25 & 2 & 6 \\
        Pick two and place & 45 & 0 & 0 \\
        \midrule
        Total & 200 & 44 & \textbf{58} \\
        \bottomrule
    \end{tabular*}
\end{table}

Each task has one valid skill and three fixed hard negatives retaining most text but corrupting a critical method, artifact, or constraint. Text Relevance uses the text reranker; + Capability Witness adds the requirement-covering Max--Min signal over this controlled structure.

Both conditions share candidates and a Qwen2.5-7B controller with deterministic decoding, a 30-step limit, and the same instruction to follow the selected skill's method, outcome, and constraint. Only ALFWorld's task-completion signal counts as success.

Paired outcomes are 18 witness-only and 4 text-only successes, 40 joint successes, and 138 joint failures (exact two-sided McNemar $p=0.004$). The controlled negatives test the witness mechanism, not unconstrained deployment.

\balance
\subsection{Feedback-Based Error Detection}
\label{app:feedback-robustness}

\paragraph{\textbf{Controlled corruption.}}
We sample 300 test queries (seed 101) and replace one query-relevant shared node per target graph with a plausible same-type cluster alternative. The 25\%, 50\%, and 75\% settings give the probability of also replacing its raw-text label, removing a clean cue. Two model seeds (42, 43) crossed with three corruption seeds (201--203) give six runs (Table~\ref{tab:graph-feedback-diagnostics-full}).

\paragraph{\textbf{Detection metrics.}}
Inserted errors are positives; nodes rank by $1-c_z$, where $c_z=r_z$ is query-independent structural confidence, not relevance. AUPRC is average precision over pooled node records per run. If $N$ nodes contain $E$ errors and reviewing the lowest-confidence $K$ finds $T$, we compute:
\begin{equation}
    \mathrm{Recall@20\%}=\frac{T}{E},\qquad
    \mathrm{Enrichment@20\%}=\frac{T/K}{E/N}.
\end{equation}
Here $K=\lceil0.2N\rceil$; boundary ties receive expected random tie-breaking contribution. Random selection has about 20\% recall and $1\times$ enrichment; random AUPRC is 0.090, near error prevalence.

Within-Graph AUC pools error--clean pairs within each query--skill graph (half credit to ties; 0.5 is random). Threshold F1 uses confidence below 0.50 to predict errors.
\begin{table}[H]
  \centering
  \footnotesize
  \caption{Error detection under raw-evidence corruption. Columns indicate corruption levels; values are mean $\pm$ SD.}
  \label{tab:graph-feedback-diagnostics-full}
  \setlength{\tabcolsep}{1pt}
    \begin{tabular*}{\columnwidth}{@{\extracolsep{\fill}}lrrr@{}}
      \toprule
      Metric & 25\% & 50\% & 75\% \\
      \midrule
      AUPRC & $0.799\pm0.015$ & $0.593\pm0.004$ & $0.369\pm0.014$ \\
      Recall@20\% & $78.3\%\pm1.9\%$ & $57.6\%\pm0.7\%$ & $37.9\%\pm1.8\%$ \\
      Enrichment@20\% & $(3.92\pm0.09)\times$ & $(2.88\pm0.04)\times$ & $(1.89\pm0.09)\times$ \\
      Within-graph AUC & $0.875\pm0.024$ & $0.738\pm0.025$ & $0.609\pm0.032$ \\
      Threshold F1 & $0.845\pm0.013$ & $0.662\pm0.012$ & $0.411\pm0.016$ \\
      \bottomrule
    \end{tabular*}
\end{table}

Despite stronger raw-evidence corruption, every setting retains enrichment above $1\times$ and within-graph AUC above 0.5.

\paragraph{\textbf{Recovery diagnostics.}}
Two diagnostic runs (seeds 42 and 43) use 500 training queries and 300 test queries (Table~\ref{tab:feedback-recovery-uncertainty}). Their 10\%, 20\%, and 30\% budgets specify the fraction of eligible shared nodes selected for error injection. These budgets differ from the probabilities used to corrupt raw labels above. Candidate-score error is the mean absolute deviation from clean-graph scores on matched query--candidate pairs; margin error uses each query's highest-positive minus highest-negative score. For either error $E$, recovery is:
\begin{equation}
    \mathrm{Recovery}(\%)=100\frac{E_{\mathrm{perturbed}}-E_{\mathrm{calibrated}}}{E_{\mathrm{perturbed}}}.
\end{equation}
Zero denotes no reduction, 100\% complete recovery, and negative values larger deviations. Negligible initial error returns zero. Enrichment uses the error-prevalence ratio above; its review budget is rounded down, with a minimum of one node.

\begin{table}[H]
  \centering
  \footnotesize
  \caption{Calibration recovery and attribution ablation, with mean $\pm$ SD across two recorded runs.}
  \label{tab:feedback-recovery-uncertainty}
  \setlength{\tabcolsep}{1pt}
  \begin{tabular*}{\columnwidth}{@{\extracolsep{\fill}}lrrrr@{}}
    \toprule
    Metric & \shortstack{Full\\10\%} & \shortstack{Full\\20\%} & \shortstack{Full\\30\%} & \shortstack{W/O Attr.\\20\%} \\
    \midrule
    Perturbed (\%) & $9.89\pm1.20$ & $20.10\pm0.98$ & $30.45\pm1.09$ & $20.10\pm0.98$ \\
    Enrich.@20\% & $3.078\pm0.180$ & $3.029\pm0.001$ & $2.944\pm0.056$ & -- \\
    Pair rec. (\%) & $34.38\pm0.85$ & $33.58\pm5.08$ & $35.38\pm4.32$ & $29.60\pm4.19$ \\
    Score rec. (\%) & $32.42\pm0.71$ & $32.45\pm4.43$ & $32.01\pm2.74$ & $29.33\pm3.16$ \\
    \bottomrule
  \end{tabular*}
\end{table}

\end{document}